\documentclass[sigconf]{acmart}

\AtBeginDocument{%
  \providecommand\BibTeX{{%
    \normalfont B\kern-0.5em{\scshape i\kern-0.25em b}\kern-0.8em\TeX}}}

\copyrightyear{2023} 
\acmYear{2023} 
\setcopyright{acmlicensed}\acmConference[DEC '23]{Second ACM Data Economy Workshop}{June 18, 2023}{Seattle, WA, USA}
\acmBooktitle{Second ACM Data Economy Workshop (DEC '23), June 18, 2023, Seattle, WA, USA}
\acmPrice{15.00}
\acmDOI{10.1145/3600046.3600051}
\acmISBN{979-8-4007-0846-6/23/06}

\usepackage{tikz}
\newcommand{\eat}[1]{}

\begin{document}


\title[Adversarial Learning in Real-World Fraud Detection: Challenges and Perspectives]{Adversarial Learning in Real-World Fraud Detection:\\ Challenges and Perspectives}

\author{Daniele Lunghi}
\authornote{The author pursues a joint PhD degree under the auspices of DEDS (No 955895), a Horizon 2020 MSCA ITN, and he is co-affiliated with Université Libre de Bruxelles in Belgium, Athena Research Center and the University of Athens (a degree awarding institute for Athena R.C.) in Greece.}
\affiliation{%
  \institution{Université Libre de Bruxelles, \\University of Athens, and Athena RC}
   \city{Bruxelles}
   \country{Belgium}
}
\email{daniele.lunghi@ulb.be}

\author{Alkis Simitsis}
\affiliation{%
  \institution{Athena Research Center}
   \city{Athens}
   \country{Greece}
}
\email{alkis@athenarc.gr}

\author{Olivier Caelen}
\affiliation{%
  \institution{Worldline S.A., Belgium}
   \city{Bruxelles}
   \country{Belgium}
}
\email{olivier.caelen@worldline.com}

\author{Gianluca Bontempi}
\affiliation{%
  \institution{Université Libre de Bruxelles}
   \city{Bruxelles}
   \country{Belgium}
}
\email{gianluca.bontempi@ulb.be}

\begin{abstract}
Data economy relies on data-driven systems and complex machine learning applications are fueled by them. Unfortunately, however, machine learning models are exposed to fraudulent activities and adversarial attacks, which threaten their security and trustworthiness. In the last decade or so, the research interest on adversarial machine learning has grown significantly, revealing how learning applications could be severely impacted by effective attacks. 
Although early results of adversarial machine learning indicate the huge potential of the approach to specific domains such as image processing, still there is a gap in both the research literature and practice regarding how to generalize adversarial techniques in other domains and applications. 
\eat{
\textcolor{blue}{Fraud detection is a particularly interesting application, due to the reciprocal influence between modern data economy and online payment systems, and for the machine learning challenges fraud detection poses, the understanding of which can help in multiple other machine learning domains.}
\textcolor{blue}{In this work we show how attacks against fraud detection systems differ}
from other applications of adversarial machine learning, 
and propose a number of interesting directions to bridge this gap.
}
Fraud detection is a critical defense mechanism for data economy, as it is for other applications as well, which poses several challenges for machine learning. In this work, we describe how attacks against fraud detection systems differ from other applications of adversarial machine learning, and propose a number of interesting directions to bridge this gap.

\end{abstract}



\maketitle

\section{Introduction} 
\label{sec:Intro}


We live in the era of a new, fourth paradigm of discovery~\cite{HTTG09} based on data-intensive science and data-driven methods and algorithms. Business decisions are increasingly based on data-driven Machine Learning (ML) algorithms and data residing in a plurality of sources (often offered by data marketplaces) and formed in various modalities. Such a modern data economy ecosystem rapidly changes how the economy works and provides immense economic and social value to our society. The emerging field of Data Economy aims at building the right tools and safeguards to ensure that the 'right' algorithms interact with the 'right' data at the 'right' price. 'Right' can be interpreted in various ways including 'fair', 'just', 'explainable', and 'secure' among others. In this study, we focus on the ‘secure’ aspect of Data Economy. As data-driven systems play a crucial role in many applications and are indispensable for many scientific, economic, and governmental activities, we should not neglect the immense risks of fraudulent activity and we need to reinforce such systems with secure learning algorithms and practices.

The necessity of operating machine learning models in adversarial environments, where an adversary actively works to have the implemented model behave in a different way from what it was defined for, led to the creation of a new research field called Adversarial Machine Learning (AML). Over the last two decades, adversarial machine learning has become a research topic of increasingly growing interest, especially due to the significant initial results obtained in the field of image recognition~\cite{papernot2017practical, shafahi2018poison}.

\eat{ 

The modern data economy ecosystem provides immense economic and social value to our society. 
Nowadays, data-driven systems play a crucial role in many applications and are indispensable for many scientific, economic, and governmental activities. 
Given the ubiquity of such systems, we should not underestimate the importance of reinforcing them with secure learning algorithms and practices.
The necessity of operating machine learning models in adversarial environments, where an adversary actively works to have the implemented model behave in a different way from what it was defined for, led to the creation of a new research field called Adversarial Machine Learning (AML). 
Over the last two decades, 
adversarial machine learning has become 
a research topic of increasingly growing interest, especially due to the 
significant initial results obtained in the field of image recognition~\cite{papernot2017practical, shafahi2018poison}.

} 

Attacks designed against an algorithm's training set (poisoning attacks) and at test time (evasion attacks) make machine learning systems highly vulnerable to attacks in a constrained domain. 

Despite the successful application of adversarial techniques to image recognition, generalizing them to other applications and domains is neither trivial nor obvious. For example, image recognition presents relatively few semantic and lexical constraints on the data, and adapting algorithms designed for it to applications where such constraints are relevant presents serious challenges~\cite{alhajjar2021adversarial, carminati2020evasion}. 
For the rest of this work, we will refer to 
such applications as `constrained applications'. 
An additional challenge is that most attacks have been developed against static systems, 
whereas many applications 
operate on streaming data. 
%
Research has only recently dealt with adversarial attacks against online systems, focusing so far mainly on the theoretical aspect of the problem~\cite{mladenovic2021online, korycki2022adversarial}.

However, this could
have substantial economic implications, as many 
valuable targets in our economy are constrained and often are online applications too. 
A case in point is bank fraud detection, which heavily relies on data-driven systems. The characteristics of the domain impose specific constraints on the data that should be taken into consideration during the design of adversarial strategies. 
For example, transactions cannot have a negative amount. Moreover, fraud detection is usually performed on aggregated features, i.e., features obtained by combining multiple transactions to observe the customers' behavioral patterns~\cite{bahnsen2016feature, leborgne2022fraud}, which
depend on the past usage of an account.
Such constraints limit the range of possible actions of an attacker.
Furthermore, the changing habits of users impose that the system continuously adapts to the environment through concept drift adaptation algorithms~\cite{DalPozzolo_2015_Credit}.

With accelerated digitalization, new risks of cybercrime are emerging. Fraudsters continuously find new ways to make financial gains, forcing the payment systems to put more and more effort into fraud detection systems~\cite{WordlineReport}. 
Global losses from payment fraud have tripled from \$ 9.84 billion  in 2011 to \$ 32.39 in 2020, an increase of more than 200\%~\cite{merchantsavvy}. 
This number could significantly increase if skillful attackers 
attempt to effectively trick 
the machine learning systems underlying the fraud detection engines.
For instance, smart fraudsters may 
attempt to understand the classifier's behavior to craft undetected frauds. To reach this goal, they could 
attempt a brute-force approach: 
first compromise 
a number of cards and then perform multiple transactions to understand the model's behavior and the characteristics of the transactions it considers genuine. While testing the model, a certain number of cards may be blocked. 
Still, if the fraudster has access to enough cards, they might bypass a non state-of-the-art fraud detection mechanism. 
%
And unfortunately, in the real-world, fraudulent attempts are much more sophisticated and at times, successful as well.
Crucially, if this behavior spreads, the whole trustworthiness of online payment systems would be jeopardized.   
In fact, trust in online payment systems is a fundamental condition to foster the growth of online services, which are a significant data source. If the risk of having online transactions hijacked becomes too high, or if the procedures to avoid fraud make the payment operations too cumbersome, the existence and profitability of the whole data economy ecosystem may be jeopardized.

\textit{Our goal and contributions}.
In this paper, we consider the problem of applying adversarial machine learning techniques to fraud detection to ensure the robustness of online transaction systems against hostile attacks. Our analysis comprises the following steps. 
\vspace{-4pt}

\begin{itemize}
  \item We describe the threats that different adversarial attacks pose to data-driven systems, and also motivate why in this work, we focus on the so-called evasion attacks (see Section~\ref{sec:Problem}).
  \item We elaborate on evasion attacks, present a taxonomy of the most important attacks described in the literature, and discuss their strengths and weaknesses. We also discuss research attempts to mitigate the risk from such attacks by developing defensive mechanisms for adversarial machine learning (see Section~\ref{sec:Evasion}). 
  \item We present 
  early solutions proposed for adversarial attacks against online (streaming) applications and fraud detection systems (see Section~\ref{sec:State}).
  \item We present critical challenges and limitations, argue that there is a gap in the literature to deal with such issues, and offer our perspectives toward future research (see Section~\ref{sec:FD}).
\end{itemize}

We present next the various types of attacks typically met in security sensitive applications.

\section{Threat Modeling in AML} \label{sec:Problem}

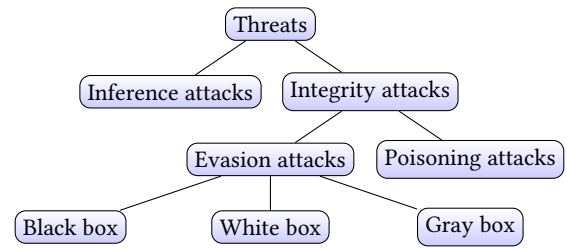
\begin{figure}[h]
    \begin{tikzpicture}[scale=0.6, sibling distance=14em,
  every node/.style = {shape=rectangle, rounded corners,
    draw, align=center,
    top color=white, bottom color=blue!20}]]
  \node {Threats}
    child { node {Inference attacks} }
    child { node {Integrity attacks}
      child { node {Evasion attacks}
        child { node {Black box} }
        child { node {White box} }
        child { node {Gray box} } }
      child { node {Poisoning attacks} }  
       };
\end{tikzpicture}
\caption{Threat model based on estimating the attacker's goals and capabilities}
\label{fig:Attacks_structure}
\vspace{-9pt}
\end{figure}

The best way to model a problem in security applications is in terms of threats, and \textit{threat modeling} is generally the first step for 
further analysis on the topic~\cite{joseph2018adversarial, carminati2020evasion}. For adversarial machine learning, this translates into modeling possible attackers based on their goals, their knowledge, and their capabilities.
Based on these axes, we provide an overview of the different fields of adversarial machine learning (see also Figure~\ref{fig:Attacks_structure}).

\textit{Inference vs. integrity attacks}.
A relevant criterion to map the threats is to consider the nature of the security violation~\cite{carminati2020evasion}.
A primary distinction is between adversaries who want to infer information from the system and those who wish to influence its behavior somehow.
The first group will perform the so-called~\textit{inference attacks}, where the goal is to obtain information about the training set of the model. Machine learning models are trained on data, and when the training data is 
sensitive, 
there is a risk that some unwanted information may leak from the behavior of the trained model. 
For instance, network models dealing with natural language texts could unintentionally memorize rare or unique sequences, which a skillful attacker may retrieve and exploit~\cite{carlini2019secret}.

Instead, influencing the system can interfere with two properties of the system: availability and integrity~\cite{carminati2020evasion}. In the former, the attacker may target the availability of certain system operations by using false positives to create Denial of Service Attacks~\cite{joseph2018adversarial}.
Instead, \textit{Integrity attacks} aim to use the false negatives to perform operations that the system is not meant to allow. In turn, integrity attacks are divided into poisoning and evasion attacks.

\textit{Poisoning attacks}.
In those attacks, the attacker may access the system's training set and inject or modify one or multiple observations to influence the model's training process. 
Such attacks are called \textit{poisoning} attacks and aim at influencing the model's behavior.
Example techniques are the backdoor attacks, 
where a backdoor key is inserted in the model. The backdoor does not affect the classifier's performance on most samples. Instead, it 
becomes active when a particular key is contained in the data, effectively allowing the attacker to infect the model unnoticed~\cite{chen2018detecting}. 
However, in fraud detection, the focus of this work, the training set is generally not available online, and it is incredibly complex even for numerous fraudsters working in parallel to inject enough transactions to affect the model's training process, given the massive number of transactions systems in production process every day.

\eat{
\textit{Evasion}.
These attacks involve cases where the attacker aims at influencing the model's behavior but has no access to its training set. 
In fraud detection, the focus of our work, the \textit{evasion} attacks pose particularly high risks and are critical for multiple reasons. 
Fraudsters aim at crafting frauds that the data-driven system will not detect. }

\textit{Evasion attacks}.
These attacks involve cases where the attacker aims at influencing the model's behavior but has no access to its training set. 
Hence, in this domain, as fraudsters aim at crafting frauds that the data-driven system will not detect, their goal is primarily to influence the model. 
Hence, \textit{evasion} attacks pose high risks and are particularly critical in fraud detection. 

\textit{White and black box attacks}.
Another dimension involves \textit{white box} and \textit{black box} attacks. 
White box attacks assume that the attacker has complete knowledge of the model's structure and weights and may exploit it to craft efficient and precise attacks. Black box attacks require no prior knowledge 
and treat the target model as a black box oracle, the structure of which can be inferred through multiple interactions. While crucial to understanding the literature on adversarial machine learning, this division may occasionally be misleading. For example, an attacker may know the model's structure but not its weights, or they may see part of the training set but not know what 
model is employed. 
%
Thus, a third category {\textit{gray box} attacks has been proposed to model this gray area. 

\eat{
Moreover, not all attacks aimed at compromising a model have the same goal. For example, an attacker may want the classifier to misclassify an observation in any direction or they may want it to classify it as a precise class. For instance, if we assume that a classifier has multiple classes that lead to an outcome unfavorable to potential attackers and one class that allows them to reach their goal, they have no interest in making the classifier predict \textit{any} wrong class. Instead, they aim precisely at the class of interest. In the second case, the attack is called \textit{targeted}, otherwise it is \textit{untargeted}.
In fraud detection, attackers typically try to inject frauds that are not recognized by the system. 
} 

\
Not all attacks aimed at compromising a model have the same goal. For example, an attacker may want the classifier to misclassify an observation in any direction or they may want it to classify it as a precise class. For instance, let us assume that a classifier has multiple classes that lead to an outcome unfavorable to potential attackers and one class that allows them to reach their goal. Then, attackers have no interest in making the classifier predict \textit{any} wrong class, and instead, they aim precisely at the class of interest. 
In the second case, the attack is called \textit{targeted}, otherwise it is \textit{untargeted}. In fraud detection, attackers typically try to inject frauds that are not recognized by the system.


\section{Evasion Attacks and Defenses} \label{sec:Evasion}

In this Section, we present evasion attacks that are particularly challenging, and present ideas 
for defense against them.

\subsection{Attacks}

\textit{White box attacks}.
White box attacks are generally the easiest to perform, and over the years, an extensive array of attacks was designed. A common idea is to formulate the evasion problem as a constrained optimization problem, where the goal is modifying an observation $x$ to generate an adversarial sample $x_{adv}$, having $x$ and  $x_{adv}$ as close as possible in the original data space. All methods described here vary in the cost metric used ($L_0$, $L_1$, $L2$, $L_\infty$ \ldots) and the formulation and optimization approaches used. 

Szegedy \textit{et al.}~\cite{szegedy2013intriguing} use the $L_1$ norm and solve it using Limited-memory Broyden Fletcher Goldfarb Shanno (L-BFGS) optimization. Fast gradient sign method (FGSM)~\cite{goodfellow2014explaining} instead uses gradient ascent to maximize the loss of the classifier.
DeepFool~\cite{moosavi2016deepfool} is an attack designed mainly for linear models. It works by iteratively generating small perturbations, determining the nearest hyperplane for an input element, and projecting it beyond this hyperplane. While linearization can be performed to extend the method to non-linear problems, the attacks can hardly be used in unconstrained domains.
Another interesting method is the Jacobian-based Saliency Map Attack (JSMA)~\cite{Papernot2015TheLimitations}, a targeted attack aimed at controlling the $L_0$ norm, hence minimizing the number of features required to perform the attack. Interestingly, controlling the $L_0$ norm allows working in domains where the attacker has access only to a set of features, which may be relevant for some applications.
Similarly, Carlini \& Wagner~\cite{carlini2017towards} propose a method that minimizes the attack's $L_0$ and the $L_1$ norm, effectively allowing to design attacks for both constrained and unconstrained domains. Moreover, such an attack was proven to break classic defenses such as distillation. Finally, Elastic-Nets~\cite{chen2018ead} propose formulating the optimization problem as an elastic network regularized optimization problem. Elastic-Nets can optimize both $L_1$ and $L2$ and have shown significant transferability of the attacks. However, the form of the optimization translates into a significantly longer computation time compared to L-BFGS.

\textit{Black box attacks}.
ZerothOrder Optimization (ZOO)~\cite{chen2017zoo} is a black box attack inspired by the C\& W attack. This method uses the logits provided by the algorithm to estimate the gradients of the classification, then optimizes through Zeroth Order Optimization. Moreover, the attack has been designed to reduce the number of queries to the model (hence avoiding direct query detection) through importance sampling, hierarchical attacks, and attack space reduction.
Another approach considers 
iterative targeted/non-targeted decision-based attacks, which do not require the logits of the target system~\cite{brendel2017decision}. Instead, the attack uses a rejection sampling algorithm to track the classifier's decision boundary and design the attacks. Using only the decisions of the classifier is the most realistic setting for many machine-learning APIs.
Similarly, the OPT attack ~\cite{cheng2018query} uses only the decisions of the classifiers as inputs. OTP uses the Randomized Gradient-Free (RGF) method to estimate the gradient at each iteration rather than the zeroth-order coordinate descent method and uses the $L_1$ and $L2$ norms to decide the size of the perturbation. 

A different technique employs 
a substitute model~\cite{papernot2016transferability}. In particular, multiple queries 
are 
used to build a model that behaves similarly to the target classifier, to then design white box attacks against the substitute model. Due to the transferability of machine learning attacks, such attacks are likely to work against the original model too.
Finally, mimicry attacks~\cite{wagner2002mimicry} may be considered a simple form of adversarial attack. The idea, developed in intrusion detection, is to generate observations that avoid detection from the system by mimicking the characteristics of normal data. Mimicry attacks can work in black and gray/white settings, but knowing the features used to evaluate users' behavior allows optimizing the attack better. 

\begin{table}[]
\centering
\caption{Taxonomy of evasion attacks in AML} 
\vspace{-6pt}
\resizebox{\columnwidth}{!}{
\begin{tabular}{|l|l|l|l|}
\hline
Attack & Attacker's knowledge & Norm                                 & Required input \\ \hline
L-BFGS & White box & $ L_2$                           & /  \\ \hline
FSGM                                          & White box            & $L_1$,                                 & /                             \\ \hline
DeepFool                                      & White box            & $L_1$, $L_2$, $L_\infty$, & /                             \\ \hline
JSMA                                          & White box            & $L_1$,                                 & /                             \\ \hline
C\&W                                          & White box            & $L_0$, $L_1$,$L_\infty$, & /                             \\ \hline
Elastic-Nets                                  & White box            & $L_1$,                                 & /                             \\ \hline
ZOO                                           & Black box            & $L_0$, $L_1$, $L_\infty$,   & Logits                        \\ \hline
Decision-Based                                & Black box            & $L_2$                                 & Decisions                     \\ \hline
OPT attack                                    & Black box            & $L_2$, $L_\infty$,       & Decisions                     \\ \hline
Substitute model                              & Black box            & /                                    & Decisions                     \\ \hline
Mimicry                                       & Black/Gray box       & /                                    & /                             \\ \hline
\end{tabular}}
\label{tab:Taxonom}
\end{table}

\subsection{Defenses}


It has been advocated that the choice of features may increase the vulnerability of a model to adversarial attacks~\cite{ilyas2019adversarial}. In particular, 
a set of highly predictive features that is brittle and incomprehensible to humans, 
could potentially 
be modified without the humans noticing it.
One approach to fix this would be to create a more robust dataset that does not contain non-robust features.

An approach designed for neural networks proposes a defense mechanism based on regularization~\cite{ross2018improving}. 
This is inspired by the classic regularization employed in training, which could be seen as a weights-regularization, to implement a new technique called input-regularization. 
The key idea is 
that by reducing the effect that small changes in the data space may have on 
the classifier's decisions, 
the impact of small changes is automatically limited, and adversaries require higher leverage on the data to breach the model.

Another defense mechanism for Deep Neural Networks (DNN) is distillation, a technique 
to train a neural network using knowledge transferred from a different
DNN. 
\cite{papernot2016distillation} proposes a distillation version using the knowledge extracted from a DNN to improve its resilience to adversarial samples. This knowledge is then used to reduce variation around the inputs, using distillation to improve the generalization capabilities of the model and, consequently, its robustness towards adversarial attacks.

A widespread defense against adversarial attacks is adversarial training; i.e., the use of adversarial samples in the training of a machine learning model. 
A form of regularization~\cite{goodfellow2014explaining}, adversarial training significantly increases the model's robustness against the attacks used in the training process. However, recent works show that training a model against multiple attacks may be cumbersome~\cite{park2020effectiveness}, and training against a type of perturbations typically does not guarantee against different types of attacks~\cite{sharma2017attacking}.
While some techniques that defend against multiple perturbations exist~\cite{tramer2019adversarial}, adversarial training is still a highly incomplete defense. Moreover, the well-known trade-off between robustness and accuracy~\cite{zhang2019theoretically} implies that all the proposed defenses have a cost, and employing them when a threat is not present may result in an unjustifiable loss of accuracy for the classifier.

\section{Existing Solutions} \label{sec:State}

In this section, we review solutions related to (a)~evasion attacks against online systems, (b)~machine learning based fraud detection, and (c)~adversarial attacks on fraud detection.

\textit{Online evasion attacks}.
Past work has studied the problem of evasion attacks against speech recognition systems~\cite{gong2019real}.
Attacks against time series data such as speech and financial time series are often performed at run time without knowing the full-time series. This is because data is one-pass, and the perturbation is added each time $t$ without access to all the elements $x_{t'}$ in the series, where $t' > t$.
This approach uses reinforcement learning to model the problem,
where the attacker bases his perturbation 
on the current status of the model. The authors propose finding the optimal policy through the Imitation Learning Strategy, where the model learns from the trajectory of a competent agent called an expert. In this case, they use state-of-the-art, non-real-time adversarial example crafting techniques as the expert.

Another work focuses on two aspects of online evasion attacks: the partial knowledge attackers have of the target model, and the irrevocability of their decision, since they operate on a transient data stream~\cite{mladenovic2021online}.
The second problem is fascinating. Generally, attacks in the literature assume that the attacker can decide which points they want to change, but in a streaming environment, the attacker must decide whether they want to launch an attack in the present moment, and the decision, when taken, is irrevocable.
The authors study a deterministic variant of the problem of online adversarial learning, where the adversary must execute $k$ successful attacks within $n$ streamed
data points, where $k << n$, and re-conduct it to the classic computer science problem, named $k$-secretary problem, where one must choose the $k$ best candidates as secretaries from a randomly ordered set of n potential candidates.
Then, they formulate a stochastic variant of the problem, which better suits the classic black box adversarial attack scenario.

\textit{Data-driven fraud detection}.
Machine learning for fraud detection is a complex and widely studied problem. 
Early works focused
on profiling the users based on the idea that the same transaction may be considered fraudulent or regular 
depending on how well it fits the habits of the person who performs it~\cite{Fawcett_1997_Data}. 
More recent works have been focusing on peculiar aspects of data distributions, such as the severe class imbalance, concept drift~\cite{DalPozzolo_2015_Credit}, verification latency~\cite{Carminati_2015_BankSealer}, and the scalability of the learning process in a streaming environment~\cite{Carcillo_2017_SCARFF}.
Notably, fraud detection is mainly performed through supervised methods~\cite{DalPozzolo2018Credit}, 
as 
unsupervised approaches struggle with covering all possible scenarios of legitimate transaction activities~\cite{krivko2010hybrid}.  
Finally, a significant challenge for fraud detection research is the lack of results sharing due to confidentiality issues~\cite{Assefa2020Generating}. 
Synthetic data generators, such as the one presented in~\cite{leborgne2022fraud}, tackle this issue and allow for controlling the environment and testing against specific challenges, such as time dependency and concept drift. Still, clearly, additional efforts are needed to make fraud detection data widely available.

\textit{Adversarial attacks on fraud detection systems}.
There are few works on adversarial attacks against fraud detection systems to date. 
El-Awady~\cite{el2021adaptive} proposes an evasion attack against fraud detection systems, showing how the problem of maximizing the revenue of an attacker who has access to a set of stolen cards may be well expressed through reinforcement learning. 

Carminati \& \textit{al}~\cite{carminati2020evasion} 
studies the problem of attacking a fraud detection system through evasion. 
The paper argues in fraud detection the attacker has only access to the raw data and not to the features used by the model, and changes in the feature space may correspond to feasible changes in the data space. 
The proposed threat model considers the attacker's goals, knowledge, and leverage.
In particular, three scenarios are considered: 
\textit{(i) White box)}, the attacker knows everything;   
\textit{(ii) Black box}, the attacker does not know the detection system and training data, but knows the previous transactions performed using the card and has access to a similar dataset to the one used by the system; 
and \textit{(iii) Gray box}, the attacker knows the model's features, the same dataset used in the black box setting, and no more.
Based on the substitute model attack~\cite{papernot2016transferability}, the paper
uses the data in the training set to train a machine learning model called Oracle, against which the attacks are then designed. Finally, it identifies two features the attacker can freely change (time and amount) and compare different strategies.
}

Another relevant contribution 
analyzes the performance of various black-box evasion attacks against 
an insurance fraud detection system. 
The work isolates four constraints: the difference between editable and non-editable features, data imbalance, designing attacks unnoticed by human investigators, and the presence of non-continuous features, and proposes various solutions to adapt existing techniques to them. 
Interestingly, the authors an open-source Python library called Adversarial Robustness Toolbox (ART)~\cite{nicolae2018adversarial}. Moreover, the experiments are performed on a real-world, publicly available German data-loans dataset~\cite{Dua:2019}, even though the considered data set is significantly smaller and less imbalanced than most bank transactions datasets~\cite{paldino2022role}.

\section{Challenges and Perspectives} \label{sec:FD}


Performing adversarial attacks against fraud detection systems is not trivial, as fraud detection presents domain-specific challenges for these attacks.
First, to perform a transaction, an attacker would need access to a stolen or cloned card. 
Since this operation comes at a cost, the attackers should be highly efficient in the number of transactions performed. 
Moreover, fraud detection systems can utilize time-dependent features to work~\cite{leborgne2022fraud}, where the past transactions of a card influence the probability of any transaction being accepted. Adversarial attacks work at aggregated features level. Hence attackers shall find the transactions that, after being processed together with the past transactions, lead to the same result obtained with standard evasion attacks~\cite{carminati2020evasion}. In general, such transactions are extremely hard or even impossible to find. 
Additionally, in real-world scenarios, several transactions' features are not observable or controllable by the attacker. For example, the average number of frauds on a terminal in the last days, used in~\cite{leborgne2022fraud}, is generally unknown to users and fraudsters alike, and cannot be hence considered in the evasion attack employed. Class imbalance can also lead to a significant loss in performance for most attacks unless adequately tackled~\cite{Cartella2021Adversarial}.

\begin{figure}[t]
 \centering
 \includegraphics[ scale=0.84]{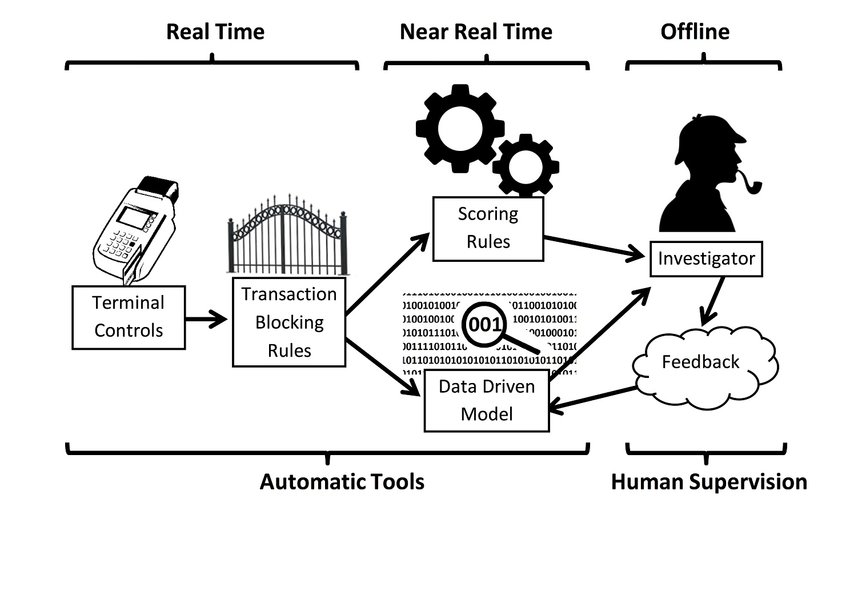}
 \vspace{-25pt}
 \caption{Fraud detection system, image taken from~\cite{Carcillo_2017_SCARFF} }
 \vspace{-15pt}
 \label{fig:fraudDetectionSystem}
\end{figure}

Furthermore, data-driven fraud detection 
employs \textit{delayed feedback}, as human investigators are often called to analyze suspicious transactions before a card is blocked~\cite{Carcillo_2017_SCARFF}
(see Figure~\ref{fig:fraudDetectionSystem}).
Hence, estimating each transaction's effect is harder for the fraudster, as the fact that a transaction is allowed does not mean that it will not lead to 
blocking the card performing it once the investigators analyze it. 
 
Finally, fraud detection systems are often performed online. As discussed in Section~\ref{sec:State}, this would require the attacks to work one-pass, and choosing the right moment to perform an attack is a further complication for the attacker.
Additionally, adaptations to concept drift may lead to continuous changes in the learner, which make it harder for a black box attacker to study it.
A common assumption in adversarial machine learning is that attacks can happen at a frequency high enough to make the drifts irrelevant. However, the problems of constrained cards budget and delayed feedback make the challenge relevant. Moreover, the frequency at which they can perform transactions with any card is limited by the automatic checks fraud detection systems perform, which result in automatic blocking of the cards. 
On the other hand, concept drift may also create opportunities for poisoning. For instance, classifier retraining means that the most recent data can disproportionately impact the model~\cite{alippi2013just}, which may mitigate the poisoning scalability issue discussed in Section \ref{sec:Problem}.

To the best of our knowledge, only a few of these challenges have been addressed in the literature. For example, 
\cite{carminati2020evasion} considers the problem of features sparsity but 
do not assume streaming settings  
or consider delayed feedback. Conversely, studies on adversarial attacks against streaming applications do not treat problems like feature observability, time dependency, or delayed feedback.

Nonetheless, this gap has significant implications 
on designing 
the right defense 
due to the difficulties of assessing the threat, which may lead to two 
conflicting challenges. First, we may
underestimate the threat, employing few or no defenses, and being vulnerable to any new effective attack crafted by the fraudsters. On the flip side, over-evaluating the risk may lead to an excessive focus on the system's robustness at the expense of accuracy. Suppose the risk is, in effect, low. 
Then, this will result in an ongoing cost in terms of accuracy, translating into an excessive number of false alarms or too many regular frauds non-detected. While this trade-off is familiar to any security application, the lack of an adequate understanding of the threats makes it significantly more dangerous.


Let us assume now that the attackers do not know the model and employ a black box attack as a substitute model attack~\cite{papernot2016transferability}. They will face a number of challenges: (a) limited number of cards they can use, (b) delayed feedback, which creates uncertainty in the information about the model each transaction provides, and (c) the fraud detection engine will treat each card differently based on their usage history. 
Even if the attackers know the transaction history of all the cards in their possession, they will still be limited in creating transactions in the feature space.
However, these challenges also depend on the budget of the attacker. For instance, if they had \textit{infinite} cards, they could use each card only once during the estimation phase, hence solving the issue of delayed feedback, and they could still be ``fast'' enough not to be affected by concept drift. They could even learn the history distribution of each card, where, assuming that each card $C$ is characterized by a history $h(p)$, they could estimate the probability distribution $P(h(C))$. 
 
Although acquiring an infinite number of cards does not seem as a viable scenario, still several critical research questions arise: ``How many cards do attackers need to pose a realistic threat?'' or ``Could they exploit other properties of fraud detection systems that we do not know to increase the efficiency of an attacks?''. Although several organizations have put together rules and policies based on experience and common logic, still a proper, principled research effort is required towards being able to assess what constitutes a real risk, much earlier than when a severe breach happens.

Admittedly, secure learning requires an accurate threat analysis, which in turn requires efforts to understand the vulnerabilities of a system and the attacks that could be performed against it. In particular, it is crucial to know how serious the threat of various families of attacks, such as poisoning and evasion, is for online and constrained applications. This would allow directing the right resources toward defense from the most likely, high-risk threats. An extra complication however is that as the fraud detection researchers and practitioners put effort in enforce their defense, at the same time, fraudsters invent novel ways to attack these defenses.

\section{Conclusions} \label{sec:Conclusions}

\eat{

Adversarial machine learning has made incredible advancements in the last decade, showing how machine learning applications are highly vulnerable to prepared and skillful attackers. Especially in computer vision, poisoning and evasion attacks have proven capable of breaching a variety of machine learning models. However, a gap between theoretical research and most real-world applications remains. First, not enough studies were made for many business-critical applications. Fraud detection, in particular, presents many challenges to an attacker significantly different from those found in image recognition. Limited budget, time-dependent features, concept drift, and verification latency present severe obstacles to existing algorithms, and research on overcoming these obstacles is still in its infancy. This leaves room for skillful attackers to create new attacks, exploiting the gap in the research and the difficulties in assessing the severity of the threat.
Finally, understanding how to deal with fraud detection challenges may help with similar applications.
First, more and more applications are deployed online, and understanding how adversarial attacks in such settings work is mandatory to deploy them in complete security. More generally, understanding how to deal with constrained applications in adversary settings is a step we must take to make the defenses and risk assessment techniques developed in the theoretical research on adversarial attacks. 

} 

Despite the recent significant advancements of adversarial machine learning, the relevant studies stress that machine learning applications are highly vulnerable to prepared and skillful attackers. In the computer vision paradigm, poisoning and evasion attacks have proven capable of breaching a variety of machine learning models. As we still have not studied adequately several business-critical applications a gap between theoretical research and many real-world applications remains.
Fraud detection, in particular, presents many challenges to an attacker that are significantly different from those studied so far in areas such as image recognition. Examples include limited budget, time-dependent features, concept drift, and verification latency. However, the same challenges also complicate the construction of effective and efficient defense mechanisms, and the relevant research is still in its infancy. 
This leaves room for skillful attackers to create new attacks, exploiting the gap in the research and the difficulties in assessing the severity of the threat.


Hence, as we have entered an era driven by a new data economy paradigm, it is imperative to fully exploit and understand how we could reinforce our data-driven, learning systems with effective yet practical fraud detection minimizing the risk of a fraudulent activity and bias. And as more and more applications operate online, research needs to adapt rapidly and present solid results towards effective defense and risk assessment, and secure operation of constrained applications in adversary settings.
Future works include comparing existing adversarial attacks against available fraud detection data sets, designing synthetic transaction generators to allow for testing adversarial attacks against concept drift, delayed feedback, and the other main challenges highlighted in this work. 

\eat{
Hence, as we enter an era driven by a new data economy paradigm, it is imperative to fully exploit and understand how we could reinforce our data-driven, learning systems with effective yet practical fraud detection. And as more and more applications operate online, research needs to adapt rapidly and present solid results towards effective defense and risk assessment, and secure operation of constrained applications in adversary settings. 
} 

\balance

  \bibliographystyle{Source/ACM-Reference-Format}
  \bibliography{Source/biblio}
\end{document}